%% file: main.tex
\documentclass[conference, letterpaper]{IEEEtran}
\usepackage{geometry}
\usepackage{graphicx}
\usepackage{amsmath}
\usepackage{amssymb}
\usepackage{multirow}
\usepackage{subfig}
\usepackage{cite}
\usepackage{hyperref}
\usepackage{color}
\usepackage{flushend}
\usepackage{booktabs}
\usepackage{tikz}
\usetikzlibrary{shapes, arrows, arrows.meta, positioning, fit, backgrounds, calc}

\usepackage{tikz}
\usetikzlibrary{arrows.meta}

\title{Learning from the Right Regions: Class-Conditioned Explanation Guidance for Medical Imaging
}

\author{Zubair Faruqui, Rahul Dubey\\
Department of Computer Science, Missouri State University\\
{Zubair10@MissouriState.edu, RahulDubey@MissouriState.edu}}

\begin{document}
\maketitle

\input{sec/0_abstract}

\input{sec/1_intro}
\input{sec/2_related_work}

\input{sec/3_new_methodology}
\input{sec/4_experiments}

\input{sec/5_results}

\input{sec/6_conclusion}


\bibliographystyle{IEEEtran}
\bibliography{main}
 
\end{document}

%% file: sec/0_abstract.tex
\begin{abstract}
Deep neural networks can achieve strong predictive performance in medical imaging while relying on regions unrelated to the predicted finding. Explanation-guided learning can address this problem, but multi-label classification introduces a key challenge: multiple abnormalities may coexist in an image, requiring each prediction to align with its corresponding spatial evidence. We propose a class-conditioned explanation-guided learning framework that constructs a separate explanation for each positive class and penalizes evidence outside its class-specific annotated regions. We instantiate the framework using Gradient-weighted Class Activation Mapping (Grad-CAM) and evaluate it on 14 abnormalities using EfficientNet-B1, ViT-B/16, ViT-B/32, and Swin-B. Explanation guidance substantially improves annotation alignment across all architectures: All-Saliency Precision increases from 0.140 to 0.625, 0.052 to 0.551, 0.060 to 0.403, and 0.095 to 0.570, respectively. These improvements are accompanied by smaller, architecture-dependent changes in classification performance. Our results show that class-conditioned explanation guidance can substantially improve where multi-label classifiers concentrate their explanatory evidence while largely preserving predictive performance.
\end{abstract}


%% file: sec/1_intro.tex
\section{Introduction}
\label{sec:introduction}

Deep neural networks have achieved strong predictive performance in medical image analysis, including the classification of abnormalities in chest radiographs~\cite{litjens2017survey,rajpurkar2017chexnet}. However, predictive performance alone does not reveal whether a model arrives at its predictions using appropriate image evidence. A classifier may exploit artifacts, acquisition-specific patterns, or other correlated features that are predictive in the training data but unrelated to the abnormality of interest~\cite{zech2018generalization,badgeley2019hipfracture}. Such reliance can lead to unexpected behavior when imaging conditions, institutions, or patient populations change. 
In medical imaging, it is therefore important not only to predict the correct abnormality, but also to determine whether the prediction is based on spatial evidence relevant to that abnormality.

Post-hoc explanation methods provide one way to examine which image regions contribute to a model's prediction~\cite{dubey}. Grad-CAM, for example, generates class-specific heatmaps that highlight regions associated with a model output~\cite{selvaraju2017gradcam}. However, these explanations are generated after training and therefore do not influence which image regions the model learns to use. Explanation-guided learning takes a different approach by using spatial information during training to encourage model explanations to align with domain-provided annotations~\cite{ross2017right,rieger2020interpretations}. In chest radiographs, radiologist-provided bounding boxes offer a practical source of such supervision by identifying regions associated with specific abnormalities without requiring pixel-level segmentation masks~\cite{li2018thoracic,nguyen2022vindr}.

Extending explanation-guided learning to multi-label medical images introduces an additional challenge. A single chest X-ray may contain multiple abnormalities, and each abnormality may be associated with a different spatial region. If their annotations are treated as a common source of spatial supervision, an explanation for one abnormality could be rewarded for focusing on a region annotated for another abnormality. Explanation supervision in the multi-label setting must therefore preserve the correspondence between each abnormality and its own spatial annotation.

To address this problem, we propose a class-conditioned explanation-guided learning framework for multi-label medical image classification. For each positive abnormality in an image, we generate a separate class-specific Grad-CAM explanation and compare it only with the spatial annotations for that abnormality. Saliency that falls outside the corresponding annotated regions is penalized during training. The class-specific penalties are first aggregated within each abnormality and then across abnormalities to form the explanation loss. This loss is combined with the conventional multi-label classification loss, encouraging each prediction to rely on spatial evidence associated with the corresponding abnormality.


We evaluate the proposed framework on VinDr-CXR~\cite{nguyen2022vindr} across 14 abnormalities using EfficientNet-B1~\cite{tan2019efficientnet}, ViT-B/16, ViT-B/32, and Swin-B. For each architecture, we examine classification performance and class-specific explanation alignment across different strengths of explanation guidance. The results show substantial improvements in explanation alignment across all four architectures, while changes in classification performance are comparatively smaller and architecture dependent. These findings indicate that explanation guidance can strongly influence where models concentrate their explanatory evidence, although its interaction with predictive performance varies across architectures. The main contributions of this work are:

\begin{itemize}
    \item We propose a class-conditioned explanation-guided learning framework for multi-label medical image classification, where the explanation for each positive class is supervised only by the spatial annotations corresponding to that class.
    
    \item We evaluate the framework on 14 abnormalities using four architectures: EfficientNet-B1, ViT-B/16, ViT-B/32, and Swin-B, and examine both classification performance and class-specific explanation alignment across different explanation-guidance strengths.
    
    \item We show that explanation guidance consistently improves explanation--annotation alignment across the evaluated architectures, while its effect on classification performance is smaller and architecture dependent.
\end{itemize}

%% file: sec/2_related_work.tex
\section{Background}
\label{sec:related_work}

Saliency-based explanation methods are widely used to examine which image regions contribute to model predictions, but previous studies have also shown important limitations in their reliability and clinical interpretation~\cite{tonekaboni2019clinicians, rudin2019stop, adebayo2018sanity, kindermans2019unreliability, saporta2022benchmarking}. These limitations motivate the use of explanations not only for post-hoc analysis, but also as a source of supervision during model training. For chest X-rays, VinDr-CXR provides radiologist annotations for 14 abnormalities, making it possible to evaluate and supervise class-specific spatial explanations in a multi-label setting~\cite{nguyen2020vinbigdata,nguyen2022vindr}.

Several approaches have incorporated spatial information into multi-label chest X-ray models. DualAttNet combines image-level and fine-grained finding attention for multi-label lesion detection~\cite{xu2024dualattnet}, while Attri-Net derives multi-label predictions from class-specific counterfactual attribution maps~\cite{sun2024attrinet}. In these approaches, spatial information is incorporated as part of the model formulation. Our focus is different: we use class-specific spatial annotations to supervise the explanations of a joint multi-label classifier during training.

In the literature, methods for incorporating explanation into model training have been investigated and discussed~\cite{faruqui2026explanation}. 
Explanation-guided learning provides a complementary approach by incorporating
human or domain knowledge directly into the training objective. Ross  constrained input gradients to discourage reliance on features identified as
irrelevant~\cite{ross2017right}, while Rieger  penalized explanations
that disagreed with prior knowledge~\cite{rieger2020interpretations}. Boyd  incorporated human saliency into model training and evaluated the
resulting models under distribution shift~\cite{boyd2023cyborg}. Grad-CAM is
particularly suitable for spatial guidance because it produces class-specific
maps over intermediate feature representations~\cite{selvaraju2017gradcam},
allowing the explanatory evidence for an individual prediction to be compared
with spatial annotations. This property has motivated its use as a training
signal in spatially supervised learning~\cite{faruqui2026explanation}.

Several studies are particularly related to this direction. CARE applies
class-specific Grad-CAM penalties inside and outside annotated regions for
boxed minority-class examples during Convolutional Neural Network (CNN) training~\cite{zhuang2019care}.
Uddin  combined Grad-CAM with Dice-based supervision in a few-shot
prototypical framework that includes VinDr-CXR~\cite{uddin2025expert}, while
Wu  used finding-specific regions derived from lung segmentation to train
CNN and Pyramid Vision Transformer configurations~\cite{wu2026localization}.
A prior binary study investigated alternative Grad-CAM target scores and
outside-region penalties using separately trained abnormality
classifiers~\cite{faruqui2026explanation}. 
In a multi-label setting, when multiple abnormalities occur in the
same image or scan, explanation supervision must preserve the correspondence between
each predicted class and its own spatial annotations. Moreover, directly
pooling all supervised image-class pairs can give greater influence to a
class simply because it is represented by more positive examples during an
optimization step. Our formulation therefore constructs a separate
class-specific explanation and annotation union for each positive
image-class pair and first aggregates supervision within each represented
class before combining it across classes.

\input{sec/fig_method_overview}

Applying the proposed formulation across different visual
architectures introduces an additional consideration: the spatial
representation used to construct Grad-CAM differs across model
families. CNNs retain explicit spatial feature maps through
convolutional layers, whereas Vision Transformers (ViTs)
represent images as sequences of patch tokens processed through
self-attention~\cite{dosovitskiy2021vit}. Swin Transformers use hierarchical
representations based on local window attention and feature
merging~\cite{liu2021swin}. Explanations for ViTs have also been studied
with respect to faithfulness and related
properties~\cite{komorowski2023evaluating,wu2024faithfulness}, and
transformer-specific attribution methods include attention
rollout~\cite{abnar2020quantifying} and relevance propagation through
attention layers~\cite{chefer2021transformer}. Together with the growing
use of transformer backbones in medical imaging~\cite{shamshad2023transformers},
these differences make the choice of spatial representation a modeling
decision rather than simply an implementation detail. Applying Grad-CAM
to these architectures requires selecting an appropriate spatial feature
representation and, for ViTs, reconstructing patch-based features into a
spatial map when necessary~\cite{wall2026gradcamvit}. We therefore
instantiate the same class-conditioned objective using convolutional,
ViT, and hierarchical transformer representations, while specifying the
architecture-specific feature and gradient paths in
Section~\ref{sec:method}.

In this work, Grad-CAM serves as
both the explanation used for spatial guidance and the primary basis for
measuring explanation--annotation alignment. Consequently, improved
localization should be interpreted as stronger alignment between the resulting
Grad-CAM explanations and class-specific annotations, rather than as direct
evidence of causal faithfulness. Against this background, we investigate how
class-conditioned spatial explanation guidance can be incorporated into a multi-label classifier and whether its effect on explanation alignment
and predictive performance persists across convolutional and transformer
architectures.

%% file: sec/fig_method_overview.tex
\begin{figure*}[t]
\centering

\resizebox{0.98\textwidth}{!}{
\begin{tikzpicture}[
    block/.style={
        rectangle,
        draw,
        rounded corners,
        align=center,
        minimum width=3.2cm,
        minimum height=1.1cm,
        font=\small
    },
    plus/.style={
        circle,
        draw,
        minimum size=0.7cm,
        inner sep=0pt,
        font=\Large
    },
    arrow/.style={
        thick,
        ->,
        >=Stealth
    }
]


\node[
    block,
    fill=blue!10
] (data) at (0,0)
{
    Multi-label Annotated Dataset\\
    (Chest X-ray)
};

\node[
    block,
    fill=green!10
] (model) at (5,0)
{
    Deep Model\\
    (CNN \& Transformer)
};


\node[
    block,
    fill=orange!10
] (prediction) at (9,0)
{
    Multi-label Disease\\
    Prediction
};

\node[
    block,
    fill=orange!10
] (celoss) at (13,0)
{
    Cross-entropy Loss\\
    $L_{\mathrm{cls}}$
};


\node[
    block,
    fill=blue!5
] (annotation) at (0,-2.0)
{
    Class-specific Expert Annotation\\
    / ROI Mask
};

\node[
    block,
    fill=red!10
] (gradcam) at (5,-2.0)
{
    Class-specific\\
    Grad-CAM Heatmap
};

\node[
    plus
] (plus1) at (8,-2.0)
{
    $+$
};

\node[
    block,
    fill=purple!10
] (exploss) at (11,-2.0)
{
    Explanation Loss\\
    Outside-Annotation\\
    Saliency Penalty\\
    $L_{\mathrm{exp}}$
};


\node[
    plus
] (plus2) at (16.5,-1.0)
{
    $+$
};

\node[
    block,
    fill=cyan!15
] (total) at (19,-1.0)
{
    Total Training Objective\\[2pt]
    $L_{\mathrm{total}}
    =
    L_{\mathrm{cls}}
    +
    \alpha L_{\mathrm{exp}}$
};


\draw[arrow]
(data.east) -- (model.west);

\draw[arrow]
(data.south) -- (annotation.north);

\draw[arrow]
(model.east) -- (prediction.west);

\draw[arrow]
(prediction.east) -- (celoss.west);

\draw[arrow]
(model.south) -- (gradcam.north);


\draw[arrow]
(gradcam.east) -- (plus1.west);

\draw[arrow]
(annotation.south)
-- ++(0,-0.8)
-| (plus1.south);

\draw[arrow]
(plus1.east) -- (exploss.west);


\draw[arrow]
(celoss.east)
-- ++(1.0,0)
|- (plus2.west);

\draw[arrow]
(exploss.east)
-- ++(3,0)
|- node[pos=0.05, left] {$\alpha$}
(plus2.west);

\draw[arrow]
(plus2.east) -- (total.west);

\end{tikzpicture}
}

\caption{
Overview of the class-conditioned explanation-aware training framework.
The model performs multi-label disease prediction and generates
class-specific Grad-CAM heatmaps. The class-specific expert annotation
is incorporated with the corresponding saliency map to compute the
explanation loss. The explanation loss is weighted by $\alpha$ and
combined with the cross-entropy loss to form the total training
objective.
}

\label{fig:explanation_aware_framework}

\end{figure*}
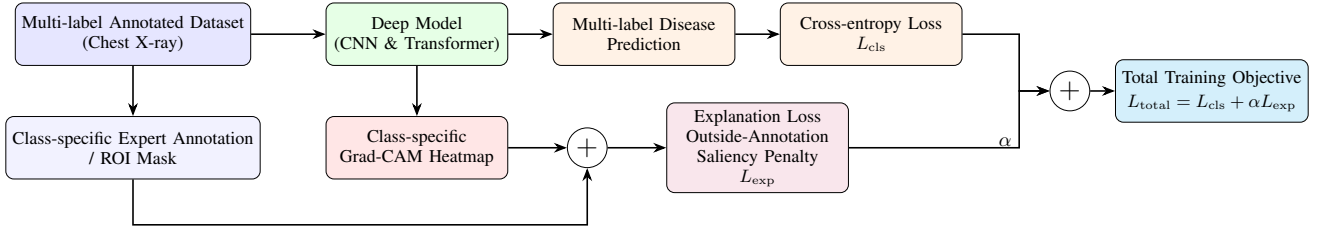

%% file: sec/3_new_methodology.tex
\section{Proposed Methodology}
\label{sec:method}

We propose a class-conditioned explanation-guided learning framework
for multi-label classification. For each positive class with an available
spatial annotation, a class-specific explanation is generated and compared
only with the annotations corresponding to that class. Explanatory evidence
outside the corresponding annotated regions is penalized, and the resulting penalties are first averaged within each represented
class and then across classes, preventing classes with more positive examples
in an optimization step from receiving proportionally greater weight in the
explanation objective.
The complete training objective combines the conventional multi-label
classification loss ($\mathcal{L}_{\mathrm{cls}}$) with the explanation loss ($\mathcal{L}_{\mathrm{exp}}$), shown in eq.~\ref{eq:overview_objective}:
\begin{equation}
    \mathcal{L}_{\mathrm{total}}
    =
    \mathcal{L}_{\mathrm{cls}}
    +
    \alpha \mathcal{L}_{\mathrm{exp}},
    \label{eq:overview_objective}
\end{equation}
where $\alpha$ controls the strength of explanation guidance.
Figure~\ref{fig:explanation_aware_framework} illustrates the overall framework.




\subsection{Problem Formulation}

Let $\mathcal{D}=\{(x_n,y_n,\mathcal{A}_n)\}_{n=1}^{N}$ denote a
multi-label training set, where $x_n$ is an input image,
$y_n\in\{0,1\}^{K}$ is its label vector, and $\mathcal{A}_n$ contains
the associated spatial annotations. For class $c$,
$y_{n,c}=1$ indicates that abnormality $c$ is present.
The spatial annotations ($\mathcal{R}_{n,c}$) associated with class $c$ are
\begin{equation}
\mathcal{R}_{n,c}
=
\{b_{n,c,m}\}_{m=1}^{J_{n,c}},
\end{equation}
where $J_{n,c}$ is the number of annotated regions for class $c$.
Importantly, annotations belonging to other classes are not included in
$\mathcal{R}_{n,c}$. This class-specific organization allows the explanation
for each predicted abnormality to be supervised independently.
A classifier $f_\theta$ produces one logit $z_{n,c}$ for each of the $K$
classes:
\begin{equation}
z_n=f_\theta(x_n), \qquad
p_{n,c}=\sigma(z_{n,c}).
\end{equation}
Because multiple abnormalities may occur simultaneously, each class
probability is modeled independently using a sigmoid output.

\subsection{Class-Specific Explanation }
\label{sec:class_conditioned_guidance}

For each ground-truth-positive image--class pair $(n,c)$ with at least one
valid spatial annotation, we construct an explanation specifically for class
$c$. This class-conditioned formulation is important in the multi-label
setting because multiple abnormalities may occur in the same image. The
explanation associated with one abnormality should therefore be supervised
using only the spatial annotations corresponding to that abnormality.

We instantiate the class-specific explanation using Grad-CAM. Let
$F_n\in\mathbb{R}^{D\times h\times w}$ denote a spatial feature tensor for
image $n$, where $D$ is the number of feature channels and $h\times w$ is the
spatial resolution. For a positive class $c$, the Grad-CAM channel weight is
computed from the corresponding raw class logit $z_{n,c}$:
\begin{equation}
\gamma_{n,d}^{(c)}
=
\frac{1}{hw}
\sum_{i=1}^{h}
\sum_{j=1}^{w}
\frac{\partial z_{n,c}}
{\partial F_{n,d,i,j}},
\qquad d=1,\ldots,D .
\label{eq:cam_weights}
\end{equation}

The resulting class-specific heatmap is
\begin{equation}
C_{n,c}(i,j)
=
\operatorname{ReLU}
\left(
\sum_{d=1}^{D}
\gamma_{n,d}^{(c)}F_{n,d,i,j}
\right).
\label{eq:class_cam}
\end{equation}

Each heatmap is independently min--max normalized:
\begin{equation}
\bar C_{n,c}(i,j)
=
\frac{
C_{n,c}(i,j)-\min_{u,v}C_{n,c}(u,v)
}{
\max_{u,v}C_{n,c}(u,v)
-\min_{u,v}C_{n,c}(u,v)+\epsilon
},
\label{eq:cam_normalization}
\end{equation}
where the minimum and maximum are computed over the $h\times w$ spatial
locations of the heatmap for the same image and class, and
$\epsilon=10^{-8}$ provides numerical stability. A separate heatmap is therefore obtained for every eligible positive class.
For an image containing multiple abnormalities, the procedure produces
multiple class-specific explanations rather than a single explanation shared
across all positive labels.


\subsection{Explanation Alignment Loss}
\label{sec:explanation_loss}
The spatial supervision for each explanation is constructed only from
annotations belonging to the corresponding class. For a positive
image--class pair $(n,c)$, all valid annotations in $\mathcal{R}_{n,c}$ are mapped
from the original image coordinate system to the preprocessed network input
using the same aspect-preserving resize and padding transformation applied to
the image. The transformed annotation boxes are then projected directly onto the native
$h\times w$ resolution of the corresponding Grad-CAM feature map. The projected annotations for class $c$ are combined into a binary union
mask $M_{n,c}\in\{0,1\}^{h\times w}$, whose entries are one at feature-grid
cells lying inside any projected class-$c$ box and zero elsewhere.

Multiple annotations for the same abnormality are therefore treated as a
single spatial region through their union. Overlapping annotation boxes remain binary
rather than being counted multiple times. Importantly, annotations belonging
to another class do not contribute to $M_{n,c}$. Thus, when several
abnormalities occur in the same image, the explanation for each class is
evaluated only against the spatial annotations associated with that class.

An annotation or bounding box marks where a finding is, but it is not an outline of it: an
axis-aligned rectangle drawn around a lesion also encloses normal tissue.
Requiring the explanation to fill the box would therefore teach the model to
spread its evidence over healthy regions. We ask only that the evidence stay
\emph{inside} the box, and leave how it is distributed there free.

Two steps implement this. First, we focus on the strongest part of the heatmap.
After normalization, we retain saliency values greater than or equal to 0.5. Annotation boxes are often much larger than the finding itself, exact contours would allow a stricter setting. Second, the penalty for one image--class pair ($\ell_{n,c}$) is the share of this retained
evidence that falls outside the class-$c$ mask, as shown in eq.~\ref{eq:nonoverlap}.
\begin{equation}
\ell_{n,c}
=
1-
\frac{\sum_{i,j}G_{n,c}(i,j)\,M_{n,c}(i,j)}
     {\sum_{i,j}G_{n,c}(i,j)}.
\label{eq:nonoverlap}
\end{equation}
It is zero when all retained evidence lies within the annotated region and
grows as more of it falls outside. The explanation loss is evaluated only for
ground-truth-positive image--class pairs that have at least one box.

\subsection{Multi-Label Explanation Aggregation}
\label{sec:represented_class_average}



The penalty $\ell_{n,c}$ is computed for every positive image--class pair
that carries an annotation. We average these penalties within each class first
and then across the represented classes, so that each abnormality contributes
equally instead of in proportion to how many annotated examples it happens to
supply. The aggregated explanation loss is computed using eq.~\ref{eq:class_balanced_exp}.
\begin{equation}
\mathcal{L}_{\mathrm{exp}}
=
\frac{1}{|\mathcal{C}|}
\sum_{c\in\mathcal{C}}
\frac{1}{|\mathcal{I}_c|}
\sum_{n\in\mathcal{I}_c}
\ell_{n,c},
\label{eq:class_balanced_exp}
\end{equation}
where $\mathcal{C}$ is the set of those classes and $\mathcal{I}_c$ their
annotated positive images. Setting $\alpha=0$ in
Eq.~\eqref{eq:overview_objective} recovers the classification-only baseline,
and the explanation term is used only during training.


The proposed objective requires a spatial feature representation for computing
Grad-CAM. For EfficientNet-B1, we use the final convolutional feature map
before global average pooling. For ViT-B/16 and ViT-B/32, the patch tokens
entering the final transformer block are reshaped to their corresponding
spatial grids after excluding the classification token. For Swin-B, we use
the final-stage spatial feature representation. 

Because Grad-CAM channel weights are themselves gradients, differentiating
the explanation objective through these weights requires second-order
derivatives. We retain this gradient path for EfficientNet-B1. For ViT-B/16,
ViT-B/32, and Swin-B, the Grad-CAM channel weights are treated as constants
during explanation-loss backpropagation. The explanation loss therefore
backpropagates through the selected spatial features but not through the
gradient computation used to obtain the Grad-CAM channel weights. In all
configurations, the classification loss follows the standard backpropagation
path.

%% file: sec/4_experiments.tex
\section{Experimental Setup}
\label{sec:experiments}

\subsection{Dataset and Preprocessing}


We use the 15,000 images from VinDr-CXR~\cite{nguyen2022vindr,
nguyen2020vinbigdata}, which carry radiologist-provided
bounding boxes for 14 abnormalities and a \emph{No finding} category. We model the 14
abnormalities and retain \emph{No finding} images as all-negative examples.
Reader annotations are pooled, and multiple boxes for the same image--class
pair are combined by union when constructing class-specific spatial masks.

The dataset is divided into 12,000 training, 1,500 validation, and 1,500 test
images, stratified by whether an image contains at least one
modeled abnormality. Images are converted to three channels and resized with
preserved aspect ratio and zero padding, with bounding boxes transformed
accordingly. Random brightness and contrast augmentation of up to 15\% is
applied during training. EfficientNet-B1 and the ViTs use
$1024\times1024$ inputs, while Swin-B uses $896\times896$ inputs.

\subsection{Training Protocol}

All models are trained with AdamW and mixed precision: EfficientNet-B1
for 50 epochs and the transformers for 30, with learning rates of
$2\times10^{-4}$ and $3$--$5\times10^{-5}$ respectively. For each model, we
train a classification-only baseline ($\alpha=0$) and explanation-guided models
with $\alpha\in\{0.1,0.2,\ldots,1.0\}$, yielding 44 configurations. The
checkpoint with the highest validation exact-match accuracy is used for test
evaluation, with a classification threshold of 0.5. Each configuration is
trained once; therefore, the reported results represent individual runs rather
than estimates of variability across repeated training.

\begin{figure}[t]
    \centering
    \includegraphics[width=\linewidth]{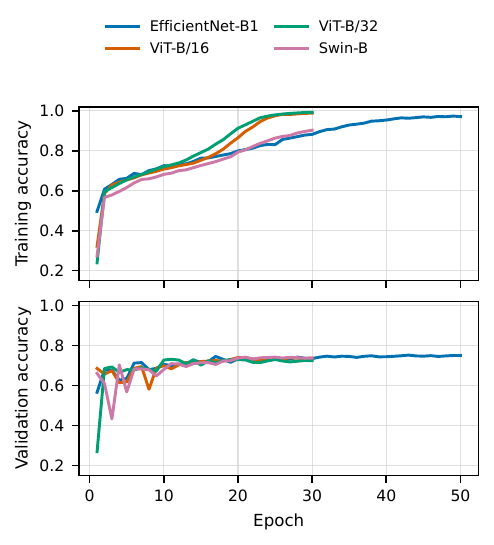}
    \caption{Training and validation accuracy of the four
    explanation-guided ($\alpha=1$) models.}
    \label{fig:training_curves}
\end{figure}

\begin{table*}[h]
\centering
\scriptsize
\setlength{\tabcolsep}{1.8pt}
\caption{Comparison of performance metrics across model architectures from $\alpha$ = 0 to $\alpha$ = 1.}
\label{tab:endpoint_results}
\begin{tabular}{lccccccccc}
\toprule
Configuration & Exact & Micro-P & Micro-R & Micro-F1 & Macro-F1 & mAP & Coverage & TSP & ASP \\
\midrule
EfficientNet-B1 & $.738\!\rightarrow\!.739$ & $.705\!\rightarrow\!.737$ & $.676\!\rightarrow\!.653$ & $.690\!\rightarrow\!.692$ & $.593\!\rightarrow\!.617$ & $.635\!\rightarrow\!.669$ & $.608\!\rightarrow\!.709$ & $.295\!\rightarrow\!.681$ & $.140\!\rightarrow\!.625$ \\
ViT-B/16 & $.715\!\rightarrow\!.736$ & $.664\!\rightarrow\!.692$ & $.576\!\rightarrow\!.619$ & $.617\!\rightarrow\!.654$ & $.453\!\rightarrow\!.519$ & $.506\!\rightarrow\!.575$ & $.539\!\rightarrow\!.548$ & $.057\!\rightarrow\!.594$ & $.052\!\rightarrow\!.551$ \\
ViT-B/32 & $.730\!\rightarrow\!.724$ & $.675\!\rightarrow\!.561$ & $.594\!\rightarrow\!.688$ & $.632\!\rightarrow\!.618$ & $.527\!\rightarrow\!.500$ & $.539\!\rightarrow\!.535$ & $.234\!\rightarrow\!.436$ & $.080\!\rightarrow\!.443$ & $.060\!\rightarrow\!.403$ \\
Swin-B & $.736\!\rightarrow\!.740$ & $.622\!\rightarrow\!.672$ & $.757\!\rightarrow\!.729$ & $.683\!\rightarrow\!.699$ & $.585\!\rightarrow\!.617$ & $.621\!\rightarrow\!.666$ & $.706\!\rightarrow\!.735$ & $.161\!\rightarrow\!.625$ & $.095\!\rightarrow\!.570$ \\
\bottomrule
\end{tabular}
\end{table*}

\begin{figure*}[h]
    \centering
    \includegraphics[width=\linewidth]{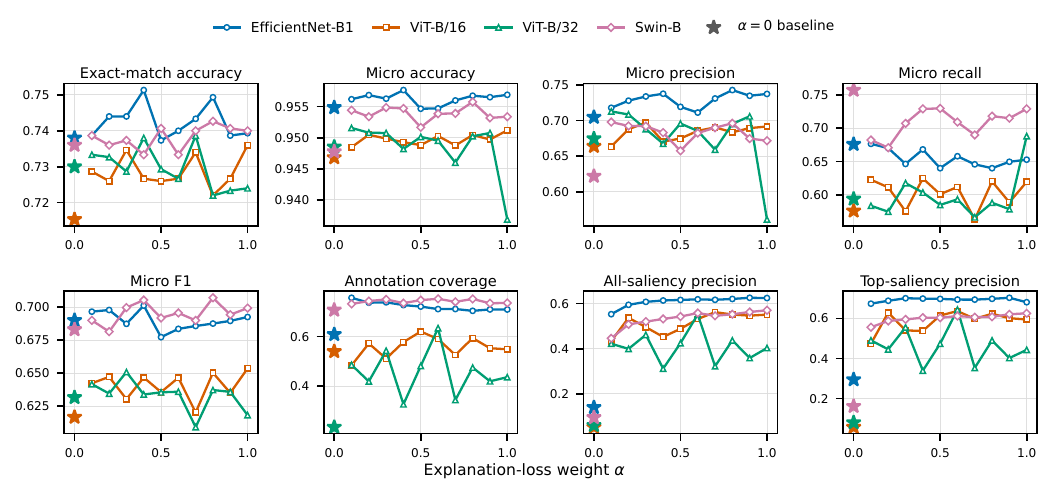}
    \caption{Cross-architecture comparison across the
    explanation-loss sweep. Lines are the guided models at
    $\alpha=0.1$--$1.0$; stars are the $\alpha=0$ baselines. Top row:
    classification. Bottom row: localization. 
    }
    \label{fig:weight_sweeps}
\end{figure*}

\begin{figure*}[h]
    \centering
    \subfloat[Annotation]{\includegraphics[width=0.158\linewidth]{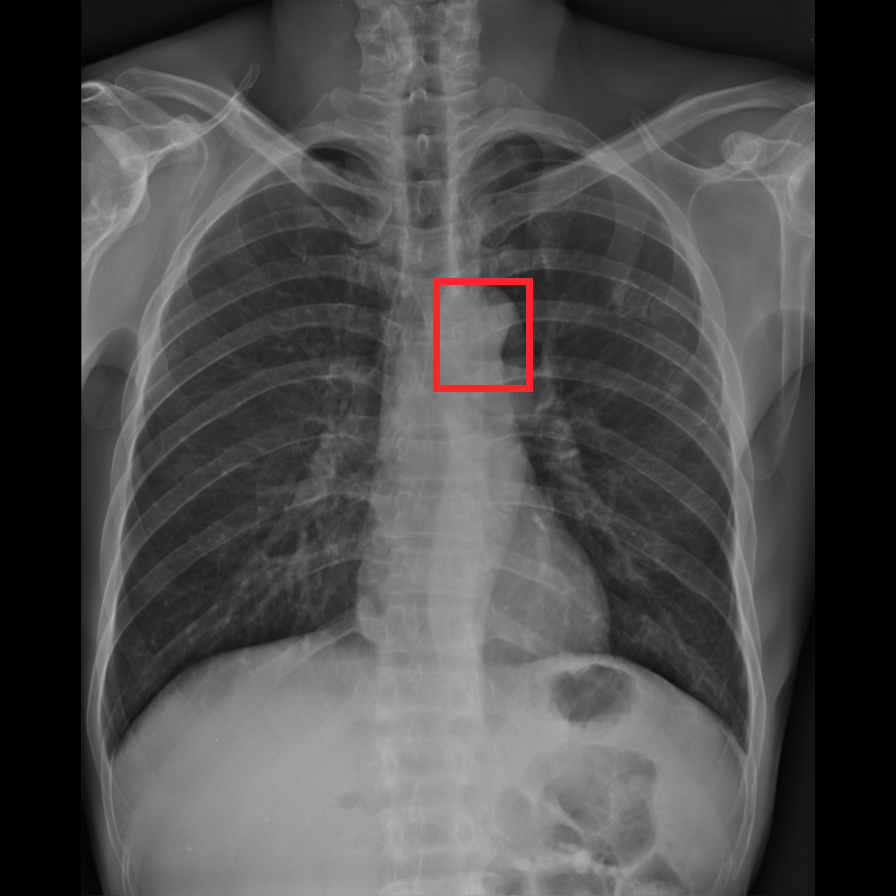}}\hfill
    \subfloat[$\alpha=0$]{\includegraphics[width=0.158\linewidth]{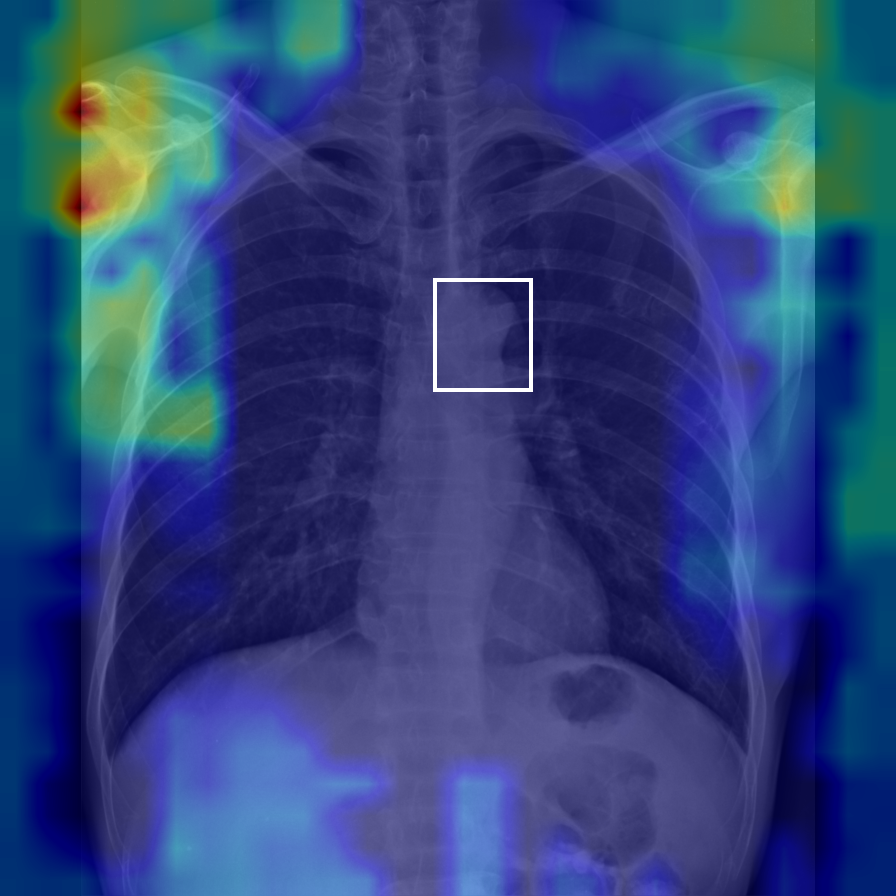}}\hfill
    \subfloat[$\alpha=0.1$]{\includegraphics[width=0.158\linewidth]{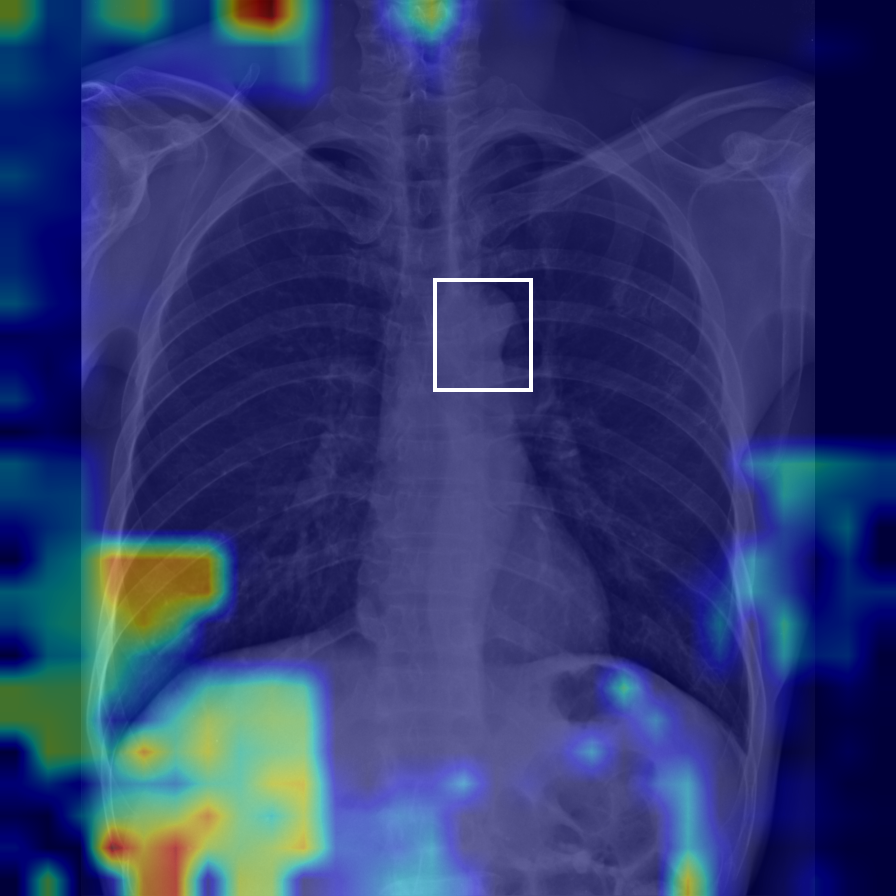}}\hfill
    \subfloat[$\alpha=0.2$]{\includegraphics[width=0.158\linewidth]{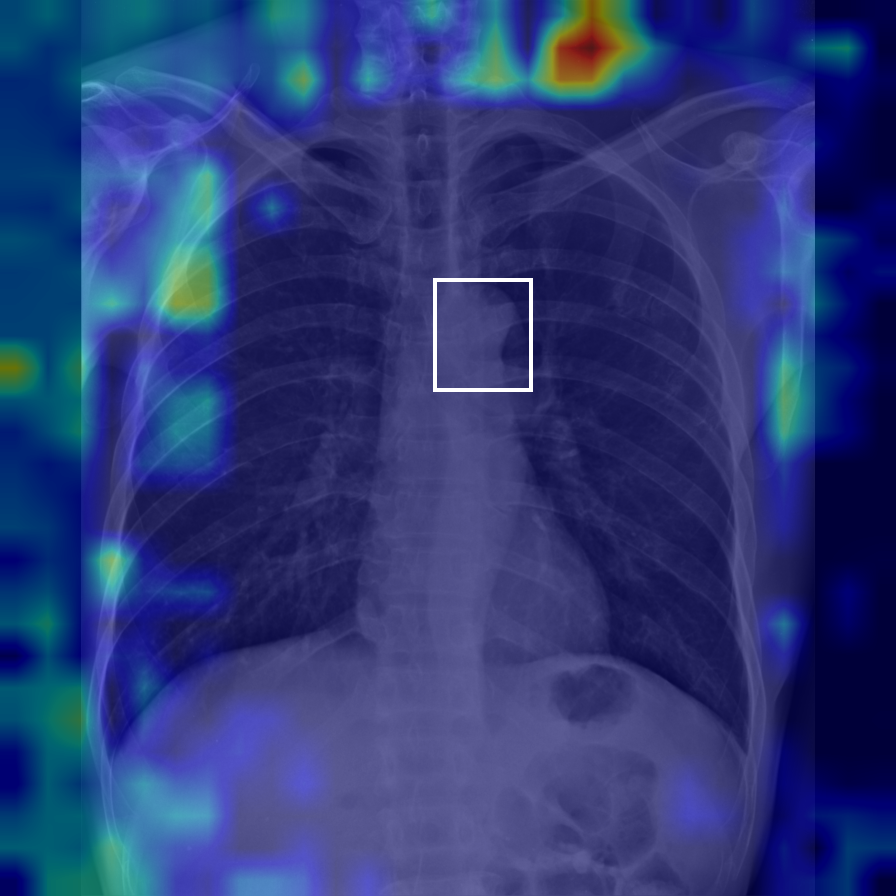}}\hfill
    \subfloat[$\alpha=0.3$]{\includegraphics[width=0.158\linewidth]{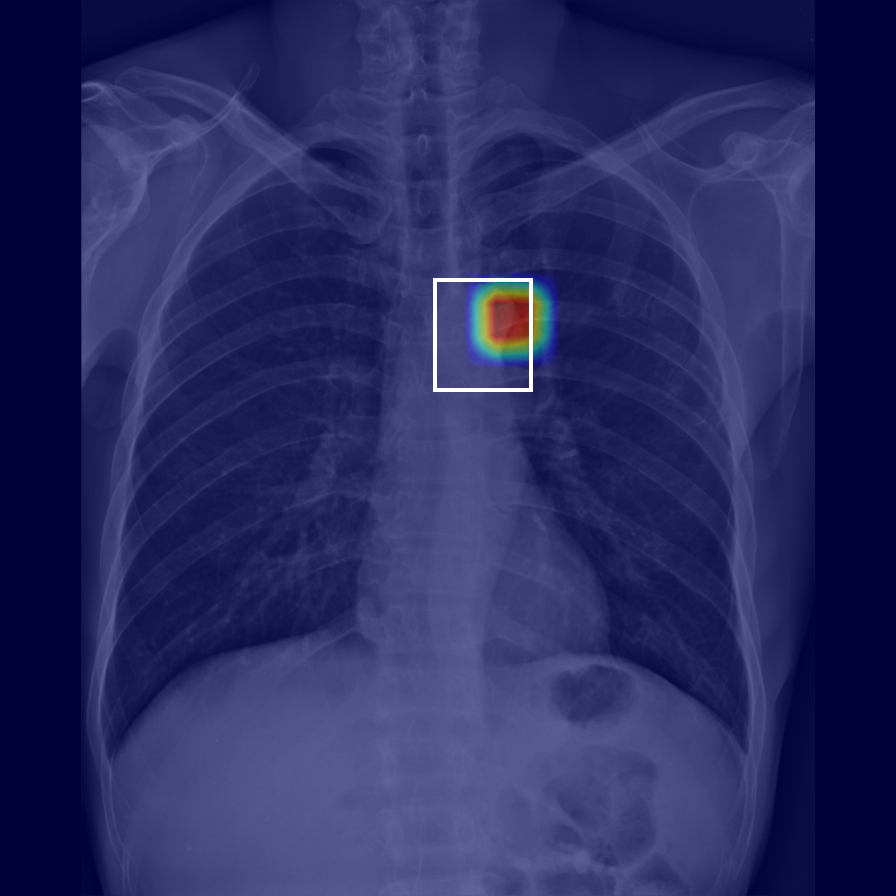}}\hfill
    \subfloat[$\alpha=0.4$]{\includegraphics[width=0.158\linewidth]{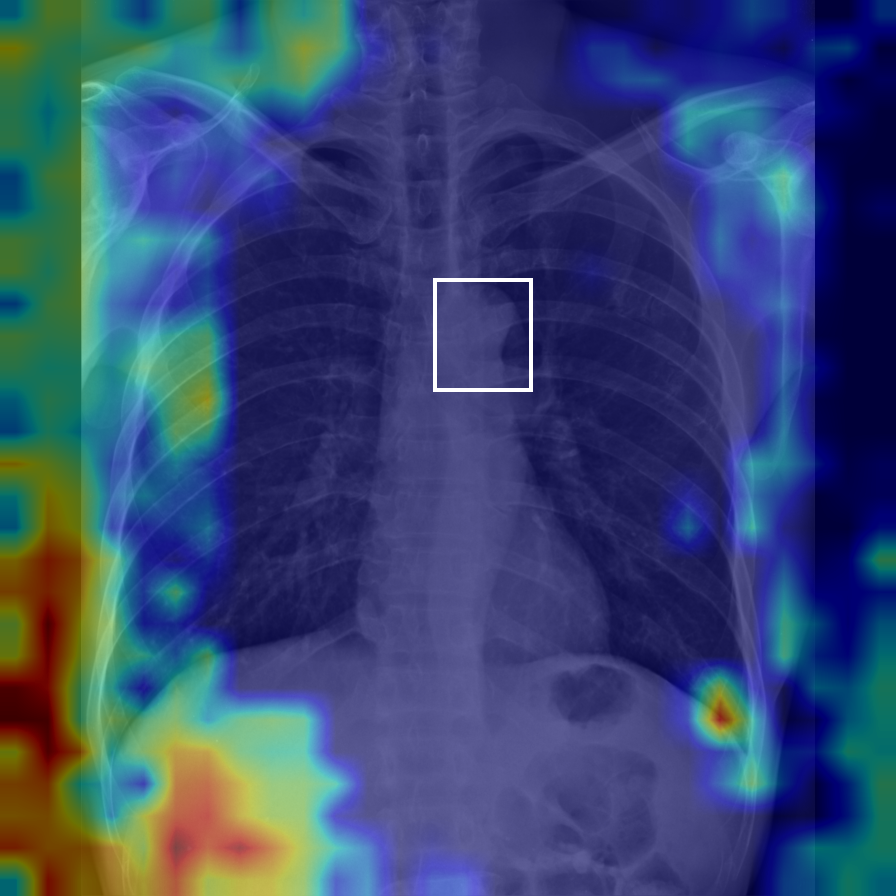}}\\[3pt]
    \subfloat[$\alpha=0.5$]{\includegraphics[width=0.158\linewidth]{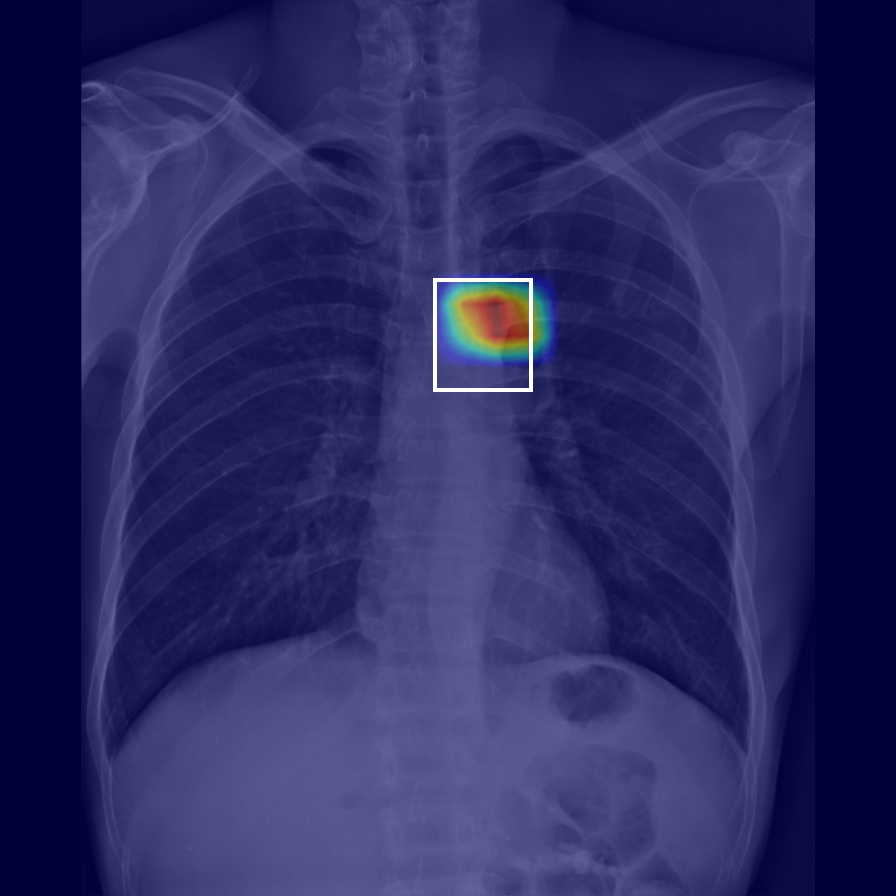}}\hfill
    \subfloat[$\alpha=0.6$]{\includegraphics[width=0.158\linewidth]{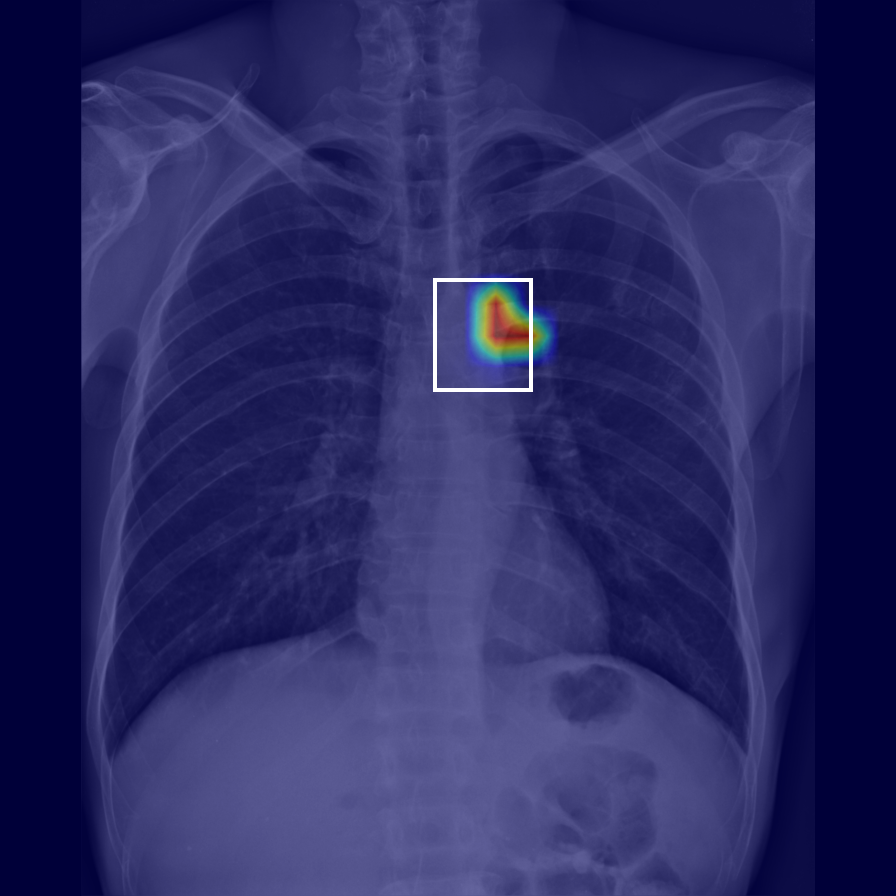}}\hfill
    \subfloat[$\alpha=0.7$]{\includegraphics[width=0.158\linewidth]{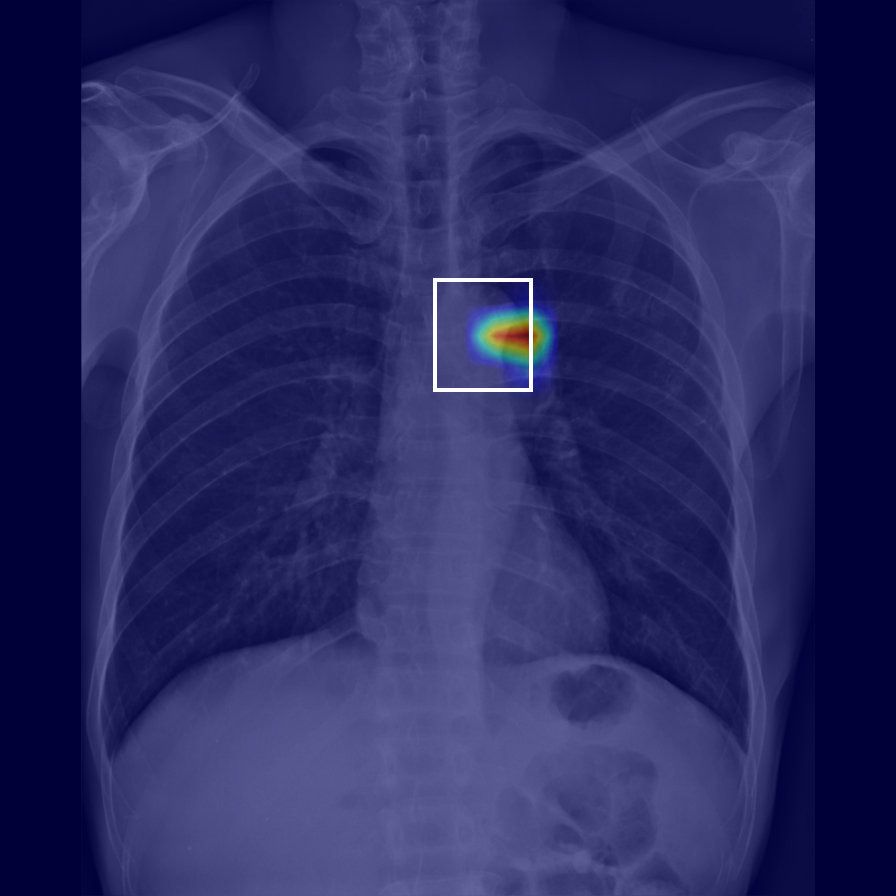}}\hfill
    \subfloat[$\alpha=0.8$]{\includegraphics[width=0.158\linewidth]{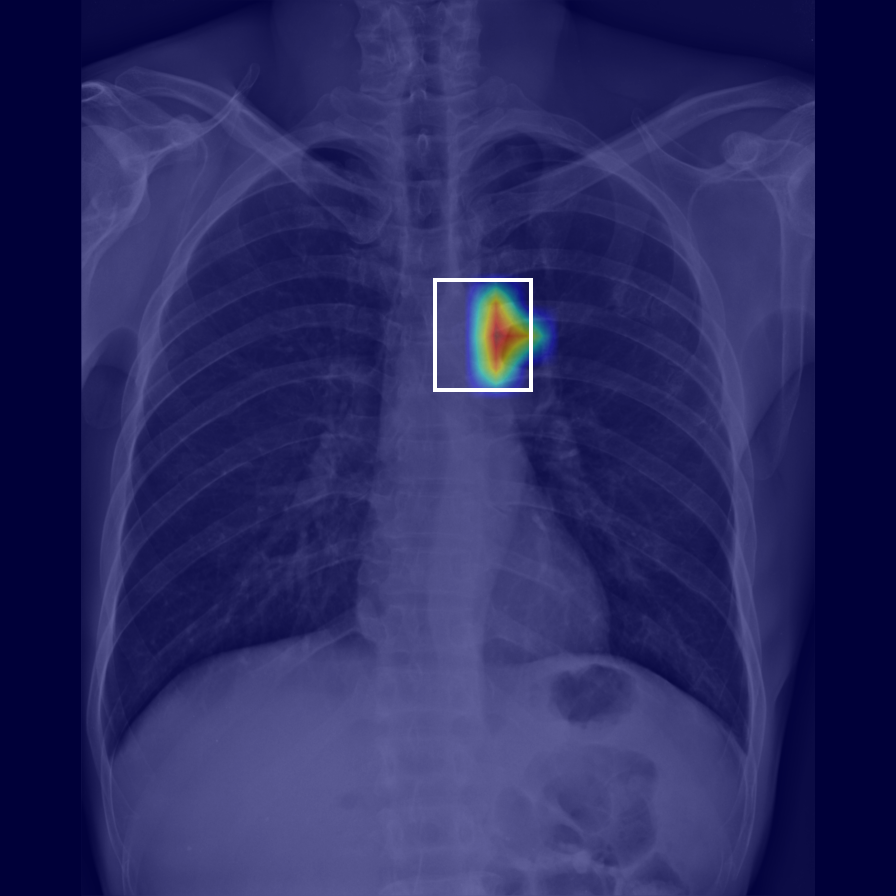}}\hfill
    \subfloat[$\alpha=0.9$]{\includegraphics[width=0.158\linewidth]{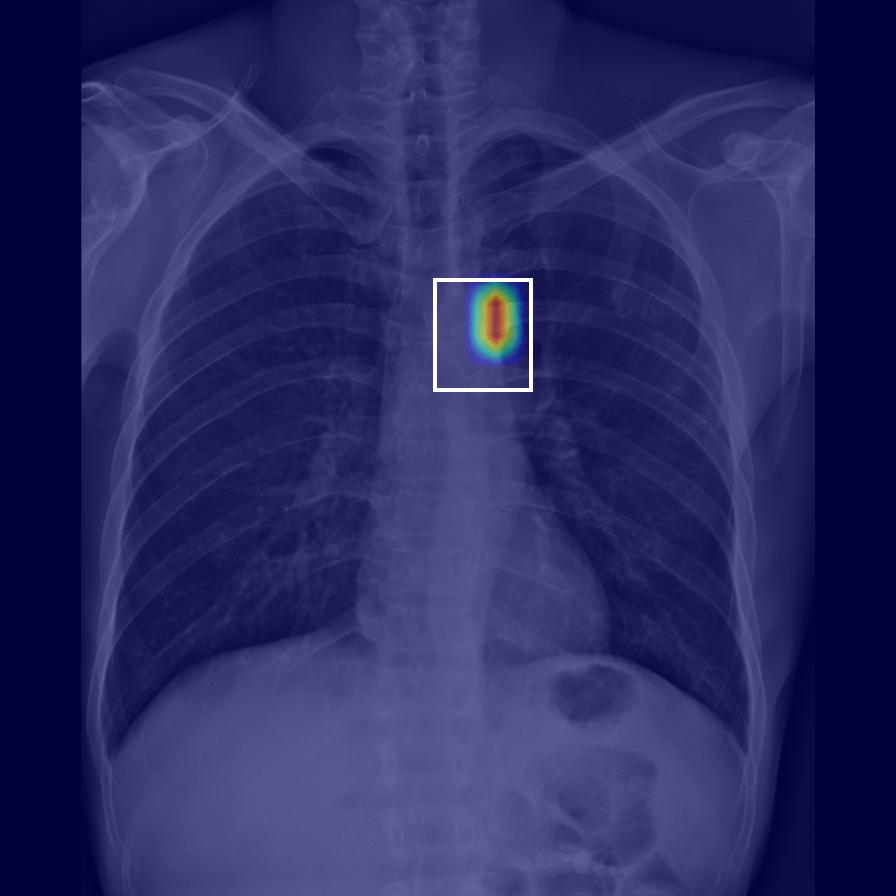}}\hfill
    \subfloat[$\alpha=1$]{\includegraphics[width=0.158\linewidth]{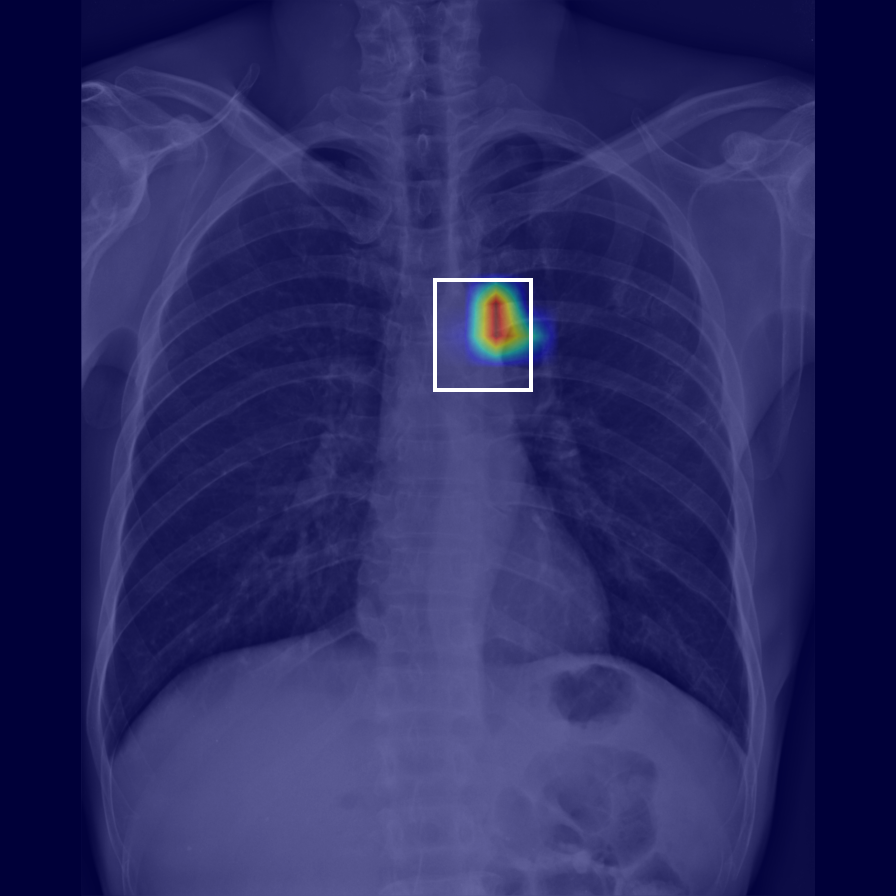}}
    \caption {Example of \textit{aortic enlargement} on one test radiograph across the eleven
    weights (Swin-B). (a) annotation; (b)--(l) Grad-CAM at $\alpha=0$ to $1$.
    Localization improves with $\alpha$, though not monotonically.}
    \label{fig:swin_examples}
\end{figure*}

\begin{figure*}[t]
    \centering
    \includegraphics[width=0.88\linewidth]{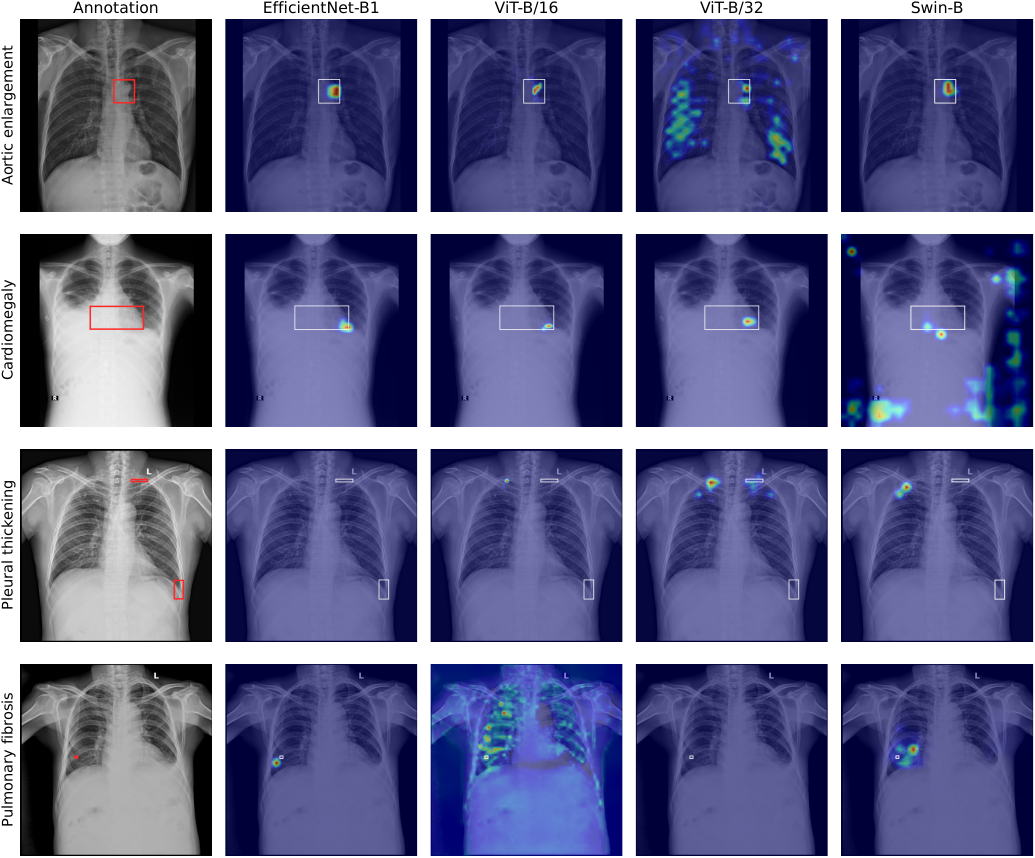}
    \caption{Guided (\(\alpha=1\)) Grad-CAM for four abnormalities on four test radiographs. Rows are the four best-supported classes; the left column shows the radiologist boxes in red, repeated in white on the model columns}
    \label{fig:cross_arch_examples}
\end{figure*}

\subsection{Evaluation Metrics}

Classification performance is evaluated on all 1,500 test images using
exact-match accuracy, micro accuracy, micro precision, micro recall,
micro-F1, macro-F1, and mean average precision (mAP). Exact match requires all
14 binary predictions for an image to be correct, while micro metrics pool
decisions across classes. Macro-F1 averages class-wise F1 scores equally, and
mAP is the unweighted mean of class-wise average precision computed from
sigmoid probabilities. Because 1,060 of the 1,500 test images are all-negative,
accuracy metrics are interpreted together with F1 and mAP.

Localization is evaluated on all ground-truth-positive image--class pairs
having at least one spatial annotation, giving 1,561 image--class pairs.
Each class-specific Grad-CAM map is upsampled to the input resolution and
independently normalized to $[0,1]$. Following~\cite{faruqui2026explanation},
we report All-Saliency Precision (ASP), which measures the fraction of total
saliency mass inside the corresponding class-specific annotation union, and
Top-Saliency Precision (TSP), which measures the fraction of saliency
thresholded at 0.5 that lies inside the annotation union. We additionally
report annotation coverage, defined as the fraction of annotation boxes for
which at least 1\% of pixels have normalized saliency of at least 0.5.

ASP and TSP are averaged over image--class pairs, whereas coverage is computed
over the 3,647 individual test-set boxes. Localization is evaluated for every
ground-truth-positive pair regardless of whether the corresponding class is
predicted as present. We compare the classification-only baseline
($\alpha=0$) with the predefined maximum guidance setting ($\alpha=1$).

%% file: sec/5_results.tex
\section{Results and discussion}
\label{sec:results}

Figure~\ref{fig:training_curves} shows the per-epoch accuracy of the
four guided models. All four fit the training set, reaching $0.90$--$0.99$
exact-match accuracy, while validation accuracy plateaus near $0.73$--$0.75$;
the checkpoint retained for testing is the validation best in every case. 

\subsection{Overall Results}

Table~\ref{tab:endpoint_results} compares the classification-only baseline
($\alpha=0$) with the maximum guidance setting ($\alpha=1$). Explanation
guidance substantially improves localization across all four configurations.
ASP increases from 0.140 to 0.625 for EfficientNet-B1, 0.052 to 0.551 for
ViT-B/16, 0.060 to 0.403 for ViT-B/32, and 0.095 to 0.570 for Swin-B.
TSP shows the same overall pattern, while annotation coverage also increases,
although by different amounts across models.

The localization improvements are accompanied by smaller and
architecture-dependent changes in classification performance. Micro-F1
changes from 0.690 to 0.692 for EfficientNet-B1, 0.617 to 0.654 for
ViT-B/16, 0.632 to 0.618 for ViT-B/32, and 0.683 to 0.699 for Swin-B.
Macro-F1 and mAP increase for three configurations but decrease slightly for
ViT-B/32. Thus, explanation guidance consistently improves Grad-CAM
alignment with the annotated regions, whereas its effect on classification
performance is mixed.

Figure~\ref{fig:weight_sweeps} shows all eight metrics across the
complete $\alpha$ sweep. Classification performance varies relatively
little across most values of $\alpha$. Micro-F1 remains within a
range of approximately 0.02--0.04 across the guided settings for
each model. Differences between architectures are generally larger
than the changes observed across most values of $\alpha$. The main
exception is ViT-B/32 at $\alpha=1$, where micro precision decreases
to 0.561, accompanied by lower micro accuracy and micro-F1.

The localization metrics show a different pattern. Much of the ASP
improvement occurs at the first nonzero guidance weight, increasing
from 0.140 to 0.553 for EfficientNet-B1, 0.052 to 0.433 for
ViT-B/16, 0.060 to 0.422 for ViT-B/32, and 0.095 to 0.446 for
Swin-B. At larger values of $\alpha$, the gains are more gradual,
reaching maximum ASP values of 0.626, 0.562, 0.552, and 0.570,
respectively. Swin-B shows a relatively smooth increase, whereas
ViT-B/16 reaches most of its improvement at low $\alpha$ and
ViT-B/32 fluctuates across the range. Coverage does not follow the
same pattern as ASP; it is already relatively high for EfficientNet-B1
and Swin-B and decreases slightly for EfficientNet-B1 at some
guidance strengths. This suggests that concentrating more saliency
inside the annotated regions does not necessarily increase the number
of annotated boxes reached by the explanation. Overall, localization
responds strongly even at low explanation-guidance weights, while
classification performance changes comparatively little across most
of the sweep.



\subsection{Class-Specific and Qualitative Results}
At the class level, ASP increases from $\alpha=0$ to
$\alpha=1$ for most abnormalities across the four models,
although this pattern is not consistent for all classes with
ViT-B/16. These class-level results should be interpreted
cautiously for abnormalities with limited test support; complete
per-class ASP, F1, and support are provided in the supplementary
material.

Figure~\ref{fig:swin_examples} follows one pair through
the whole sweep for aortic enlargement using Swin-B architecture. Raising $\alpha$
pulls the evidence out of the surrounding thorax and into the annotated region.
As shown in Table~\ref{tab:endpoint_results}, 
at \(\alpha=1\), EfficientNet-B1 achieves the highest ASP (0.625) and TSP (0.681). Among the transformer models, Swin-B achieves the highest ASP (0.570), coverage, and micro-F1.
Figure~\ref{fig:cross_arch_examples} shows four abnormalities across the four guided models, and 
guidance helps least where
the box is far smaller than a Grad-CAM cell, as in the pulmonary-fibrosis and
pleural-thickening rows, and where activation leaks onto the zero-padded border
of the letterboxed input, which is the failure visible for Swin-B on
cardiomegaly.



Finally, the spatial annotations are pooled across readers, and both training
and localization evaluation use Grad-CAM. Improvements in ASP and TSP
therefore establish stronger alignment of these Grad-CAM explanations with
class-specific annotations, but do not by themselves establish causal
faithfulness, agreement with other attribution methods, or clinical validity.
As Figure~\ref{fig:cross_arch_examples} shows, ASP is also sensitive to
annotation scale: for abnormalities annotated with very small boxes it can
report near-zero alignment for explanations that are visibly close to the
correct structure, so per-class ASP should not be compared across classes whose
boxes differ greatly in size. 

More broadly, these results show that explanation-guided learning can
be extended to the multi-label setting without treating spatial
supervision as a single annotation shared across all predictions.
Preserving the correspondence between each abnormality and its own
spatial annotations allows multiple class-specific explanations to be
guided within the same image, while retaining a single joint
multi-label classifier. The results across convolutional and
transformer architectures further suggest that this formulation is not
tied to a particular model family, although the spatial representation
used to construct the explanation remains architecture dependent.
This provides a practical basis for applying class-conditioned
explanation guidance to other multi-label medical imaging problems
where class-specific spatial annotations are available.

%% file: sec/6_conclusion.tex
\section{Conclusion}
\label{sec:conclusion}

We presented a class-conditioned explanation-guided learning framework for
multi-label chest X-ray classification, in which each positive class is guided
by its corresponding spatial annotations. Across EfficientNet-B1, ViT-B/16,
ViT-B/32, and Swin-B, explanation guidance substantially improved Grad-CAM
alignment with annotated regions, with ASP increasing by 0.343--0.499 from
$\alpha=0$ to $\alpha=1$. These gains were accompanied by smaller,
architecture-dependent changes in classification performance, with micro-F1
increasing in three of the four configurations. The results demonstrate that
class-conditioned explanation guidance can improve where multi-label models
concentrate their explanatory evidence while largely preserving predictive
performance. Future work will evaluate repeated runs, additional datasets and
attribution methods, and the clinical relevance and generalizability of the
resulting explanations.